\documentclass[final]{cvpr}
\usepackage{times}
\usepackage{epsfig}
\usepackage{graphicx}
\usepackage{amsmath}
\usepackage{amssymb}
\usepackage{physics}
\usepackage{enumitem}
\usepackage{bm}
\usepackage{microtype}
\usepackage{multirow}

\usepackage[pagebackref,breaklinks,colorlinks]{hyperref}

\usepackage{bbm}
\usepackage{caption}

\def\cvprPaperID{***} % *** Enter the CVPR Paper ID here
\def\confName{CVPR}
\def\confYear{2023}
\usepackage{pifont}% http://ctan.org/pkg/pifont
\newcommand{\cmark}{\ding{51}}%
\newcommand{\xmark}{\ding{55}}%

\begin{document}

%%%%%%%%% TITL
\title{SceneComposer: Any-Level Semantic Image Synthesis}
% \title{SceneComposer: Image Synthesis from Any-Level Semantic Layout}

% \author{Yu Zeng\\
% Johns Hopkins University\\
% % Institution1 address\\
% {\tt\small yzeng22@jhu.edu}
% % For a paper whose authors are all at the same institution,
% % omit the following lines up until the closing ``}''.
% % Additional authors and addresses can be added with ``\and'',
% % just like the second author.
% % To save space, use either the email address or home page, not both
% \and
% Zhe Lin\\
% Adobe Research\\
% % First line of institution2 address\\
% {\tt\small zlin@adobe.com}
% \and
% Vishal M. Patel\\
% Johns Hopkins University\\
% % Institution1 address\\
% {\tt\small vpatel36@jhu.edu}
% }

\author{{Yu Zeng$^1$,
\hfill
Zhe Lin$^2$,
\hfill
Jianming Zhang$^2$,
\hfill
Qing Liu$^2$,
\hfill
John Collomosse$^2$,
\hfill
Jason Kuen$^2$,
\hfill
Vishal M. Patel$^1$}\\
{\tt\small \{yzeng22,vpatel36\}@jhu.edu, \{zlin,jianmzha,qingl,collomos,kuen\}@adobe.com}\\
{\hfill $^1$Johns Hopkins University\hfill\hfill $^2$Adobe Research\hfill\hfill}\\
}

%%%%%%%%% TITLE
% \maketitle
\twocolumn[{%
\renewcommand\twocolumn[1][]{#1}%
\maketitle
\begin{center}
    \centering
    \vspace{-0.5cm}
    \includegraphics[width=\textwidth]{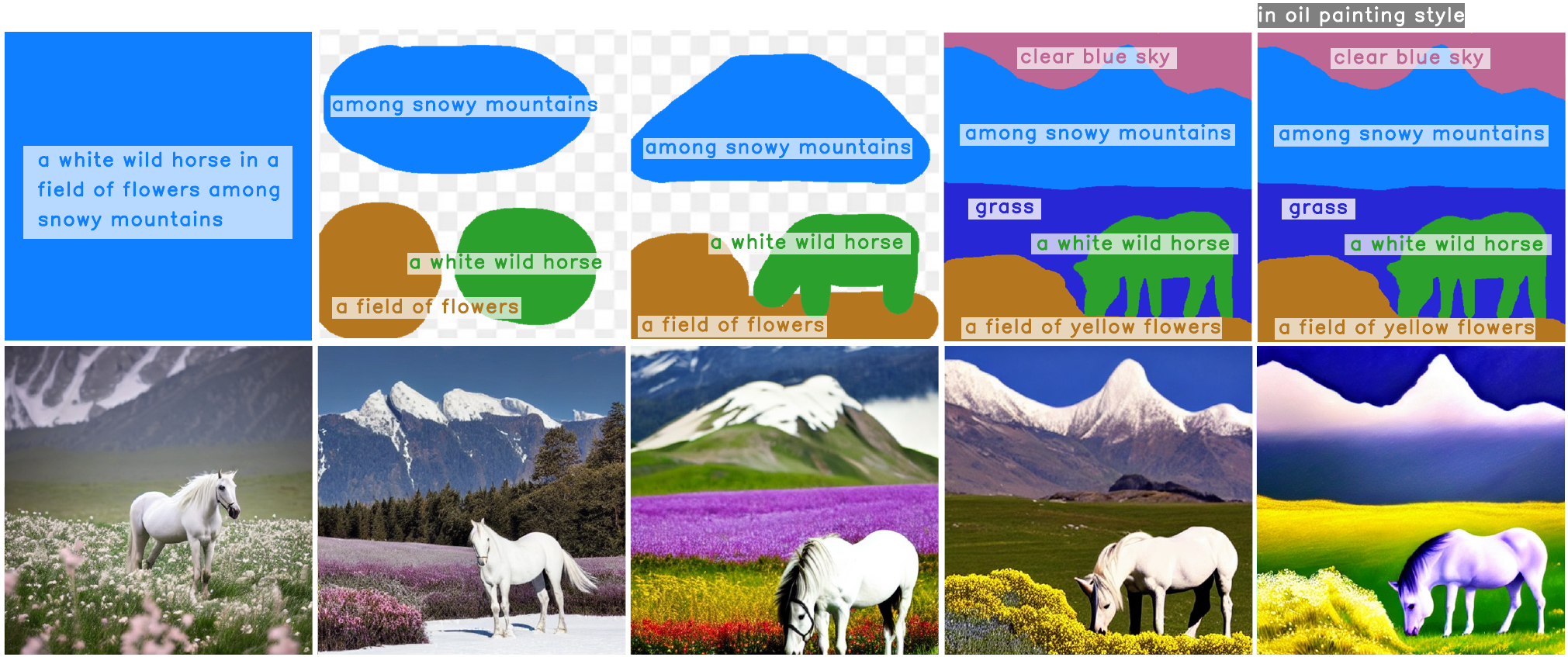}\\
    \scriptsize{\hfill {(a)} \hfill\hfill {(b)} \hfill\hfill {(c)} \hfill\hfill {(d)} \hfill\hfill {(e)} \hfill}
    % \vspace{0.30cm}
    \captionof{figure}
    {Examples of image synthesis from any-level semantic layouts.
    (a) The coarsest layout,~\ie at the $0$-th precision level, is equivalent to a text input; (d) the finest layout,~\ie at the highest level, is close to an accurate segmentation map; (a)-(d) intermediate level layouts (from coarse to fine), the shape control becomes tighter with increasing levels.
    (e) We can specify different precision levels for different components,~\eg to include a $0$-th level style indicator while the remaining regions are of higher levels. 
    }
    \vspace{-0.2em}
        \label{fig0}
\end{center}%
}] 

\makeatother

%%%%%%%%% ABSTRACT

\begin{abstract}
We propose a new framework for conditional image synthesis from semantic layouts of any precision levels, ranging from pure text to a 2D semantic canvas with precise shapes. More specifically, the input layout consists of one or more semantic regions with free-form text descriptions and adjustable precision levels, which can be set based on the desired controllability. The framework naturally reduces to text-to-image (T2I) at the lowest level with no shape information, and it becomes segmentation-to-image (S2I) at the highest level. By supporting the levels in-between, our framework is flexible in assisting users of different drawing expertise and at different stages of their creative workflow. We introduce several novel techniques to address the challenges coming with this new setup, including a pipeline for collecting training data; a precision-encoded mask pyramid and a text feature map representation to jointly encode precision level, semantics, and composition information; and a multi-scale guided diffusion model to synthesize images. To evaluate the proposed method, we collect a test dataset containing user-drawn layouts with diverse scenes and styles. Experimental results show that the proposed method can generate high-quality images following the layout at given precision, and compares favorably against existing methods. Project page \url{https://zengxianyu.github.io/scenec/}
\end{abstract}
\vspace{-0.3cm}

%%%%%%%%% BODY TEXT
\section{Introduction}
Recently, deep generative models such as StyleGAN \cite{karras2019style,karras2020analyzing} and diffusion models \cite{dhariwal2021diffusion,ho2020denoising,song2020denoising} have made a significant breakthrough in generating high-quality images.  Image generation and editing technologies enabled by these models have become highly appealing to artists and designers by helping their creative workflows. 
To make image generation more controllable, researchers have put a lot of effort into conditional image synthesis and introduced models using various types and levels of semantic input such as object categories, text prompts, and segmentation maps~\etc
~\cite{rombach2022high,nichol2021glide,saharia2022photorealistic,park2019semantic,zhu2020sean,isola2017image}. 

However, existing models are not flexible enough to support the full creative workflow. 
They mostly consider fixed-level semantics as the input,~\eg image-level text descriptions in text-to-image generation (T2I)~\cite{saharia2022photorealistic,ramesh2022hierarchical,nichol2021glide,reed2016generative,rombach2022high}, or pixel-level segmentation maps in segmentation-to-image generation (S2I)\cite{park2019semantic,zhu2020sean,isola2017image}. 
Recent breakthroughs on T2I such as DALLE2 \cite{ramesh2022hierarchical} and  StableDiffusion~\cite{stablediffusion,rombach2022high} demonstrate extraordinary capabilities of generating high-quality results. 
They can convert a rough idea into visual messages to provide inspirations at the beginning of the creative process, but provide no further control over image composition. 
On the other hand, S2I allows users to precisely control the image composition. 
As it is extremely challenging to draw a detailed layout directly, S2I is more useful for later creative stages given initial designs. 
For real-world use cases, it is highly desirable to have a model which can generate images from not only pure text or segmentation maps, but also intermediate-level layouts with coarse shapes. 

To this end, we propose a new unified conditional image synthesis framework to generate images from a semantic layout at any combination of precision levels.
It is inspired by the typical coarse-to-fine workflow of artists and designers: they first start from an idea, which can be expressed as a text prompt or a set of concepts (Fig.~\ref{fig0}~(a)), then tend to draw the approximate outlines and refine each object (Fig.~\ref{fig0}~(a)-(d)). 
More specifically, we model a semantic layout as a set of semantic regions with free-form texts descriptions. The layout can be sparse and each region can have a precision level to control how well the generated object should fit to the specified shape. 
The framework reduces to T2I when the layout is the coarsest (Fig.~\ref{fig0}~(a)), and it becomes S2I when the layout is a segmentation map (Fig.~\ref{fig0}~(d)). 
By adjusting the precision level, users can achieve their desired controllability (Fig.~\ref{fig0}~(a)-(d)). This framework is different from the existing works in many aspects, as summarized in Table~\ref{tab_intro}. 
\begin{table}[!t]
\setlength{\tabcolsep}{2pt}
\caption{\small Difference from related conditional image synthesis works. T2I: text to image, S2I: segmentation to image, ST2I: Scene-based text to image~\cite{gafni2022make}, Box2I: bounding box layout to image~\cite{sylvain2021object}. }
\begin{center}
\vskip-13pt\small
\resizebox{\columnwidth}{!}{
\begin{tabular}{cccccc}
\hline
Setting & Open-domain layout & Shape control & Sparse layout & Coarse shape & Level control\\
\hline
T2I & \cmark & \xmark & \xmark &\xmark &\xmark\\
S2I & \xmark & \cmark & \xmark& \xmark &\xmark\\
ST2I & \xmark & \cmark & \xmark & \xmark &\xmark\\
Box2I & \xmark & \xmark & \cmark& \xmark &\xmark\\
Ours & \cmark & \cmark & \cmark& \cmark &\cmark\\
\hline
\end{tabular}}
\label{tab_intro}
\end{center}
\vspace{-25pt}
\end{table}

This new setup comes with several challenges. 
First, it is non-trivial to encode open-domain layouts in image synthesis frameworks. 
Second, to handle hand-drawn layouts of varying precision, we need an effective and robust way to inject the precision information into the layout encoding. 
Third, there is no large-scale open-domain layout/image dataset. To generate high-quality images and generalize to novel concepts, a large and diverse training dataset is crucial.

We introduce several novel ideas to address these challenges. 
First, we propose a text feature map representation for encoding a semantic layout. It can be seen as a spatial extension of text embedding or generalization of segmentation masks from binary to continuous space. 
Second, we introduce a precision-encoded mask pyramid to model layout precision. Inspired by the classical image pyramid models~\cite{adelson1984pyramid,burt1987laplacian,shocher2020semantic,xu2021generative}, we relate shape precision to levels in a pyramid representation and encode precision by dropping out regions of lower precision levels. In other words, the $l$-th level of the mask pyramid is a sub-layout (subset of regions) consisting of semantic regions with precision level no less than $l$. 
By creating a text feature map for each sub-layout, we obtain a text feature pyramid as a unified representation of semantics,  composition, and precision. Finally, we feed the text feature pyramid to a multi-scale guided diffusion model to generate images. 
We fulfill the need for training data by collecting them from two sources: (1) large-scale image-text pairs; (2) a relatively small pseudo layout/image dataset using text-based object detection and segmentation. With this multi-source training strategy, both text-to-image and layout-to-image can benefit from each other synergistically. 

Our contributions are summarized as follows:
\begin{itemize}[noitemsep]
    \item A unified framework for diffusion-based image synthesis from semantic layouts with any combination of precision control.
    \item Novel ideas to build the model, including precision-encoded mask pyramid and pyramid text feature map representation, and multi-scale guided diffusion model, and training with multi-source data.
    \item A new real-world user-drawn layout dataset and extensive experiments showing the effectiveness of our model for text-to-image and layout-to-image generation with precision control.
\end{itemize}

\section{Related Work}
\noindent\textbf{Deep generative models. } In recent years, there has been significant progress in image generation using deep generative models. 
Some of these approaches attempt to learn the image distribution by optimizing a likelihood-based objective function. 
Autoregressive models  (ARMs)~\cite{van2016conditional,van2016pixel,van2017neural,esser2021taming} treat images as sequences of pixels or tokens in a learned dictionary to define a tractable density model, which can be optimized by maximizing the likelihood. 
Variational Autoencoders (VAEs)~\cite{kingmaauto,vahdat2020nvae} use an intractable density function and train the model by maximizing the variational lower bound. 
Diffusion-based models~\cite{sohl2015deep,dhariwal2021diffusion,ho2020denoising,song2020denoising,rombach2022high} are also trained by optimizing the variational lower bound.  Unlike VAEs which map an image into the latent space using a learnable encoder, diffusion-based models use a fixed diffusion process to transform an image into Gaussian noise. 
Generative Adversarial Networks (GANs)  \cite{goodfellow2014generative,brock2018large,karras2019style,karras2020analyzing} do not define the density function explicitly; instead, generative models are trained through an adversarial learning process against a discriminator. 
We base our method on diffusion models due to their ability to generate high-quality images and stability in training. 

\noindent\textbf{Layout/Segmentation-to-image generation } refers to the task of generating images from a spatial arrangement of semantic concepts,~\eg a segmentation map or a set of bounding boxes.  
Segmentation-based methods (\eg Pix2Pix~\cite{isola2017image}, Pix2PixHD~\cite{wang2018high}, CRN~\cite{chen2017photographic}, SPADE~\cite{park2019semantic}) have been proposed to allow users to precisely control the image composition, but are less flexible as they require a dense and accurate segmentation map and only allow a fixed set of categories. 
Bounding box-based methods can generate images from a coarse layout of bounding boxes. Bounding box-based generation was studied as a standalone task first by Zhao~\etal\cite{zhao2019image} and has been an active research area~\cite{he2021context,wang2022interactive,ashual2019specifying} since then. To convey more information than object categories, more expressive bounding box descriptions have been introduced, such as  attributes~\cite{maattribute,ashual2019specifying},  relations~\cite{ashual2019specifying}, and free-form  descriptions~\cite{10.1007/978-3-031-15931-2_33}. 
Although bounding boxes are easier to draw than segmentation maps, they provide no control over object shapes and orientations. In this work, we use coarse shapes with free-form text descriptions and precision levels to represent a layout, providing more flexibility and control for different synthesis needs.  

\noindent\textbf{Text-to-image/Multi-modal image generation. } T2I aims to generate images from a free-form text description,~\ie image captions. 
Earlier approaches are usually based on conditional GANs~\cite{reed2016generative,zhang2017stackgan}. Recently, autoregressive models based on image quantization techniques~\cite{ramesh2021zero,yu2022scaling,ding2021cogview} and diffusion models~\cite{rombach2022high,saharia2022photorealistic,ramesh2022hierarchical,stablediffusion,nichol2021glide} have shown surprising results with large-scale models and datasets. To gain more control over the spatial composition, some studies have tried to incorporate layout in the form of key  points~\cite{reed2016learning}, bounding boxes~\cite{hinz2018generating}, or object shapes~\cite{li2019object,hong2018inferring,li2020image,gafni2022make,pavllo2020controlling}, as an additional input. Some studies focused on image synthesis from multi-modal input including text, segmentation maps, edge maps~\cite{huang2022multimodal,wu2022nuwa}~\etc. 
A work~\cite{balaji2022ediffi} concurrent to ours proposes a  training-free paint-with-word method that enables users to specify the object locations by manipulating the attention matrices in cross-attention layers. 
These approaches use the layout as an extra signal complementary to captions. In contrast, we unify image captions and layout using a multi-level framework that models captions as a special case of layout with the lowest precision level. From a practical perspective, our framework provides a more flexible control mechanism with coarse shapes and precision controllability. 
\begin{figure*}[t]
\begin{center}
% \vspace{-10pt}
% \fbox{\rule{0pt}{1.5in} \rule{\linewidth}{0pt}}
\includegraphics[width=\linewidth]{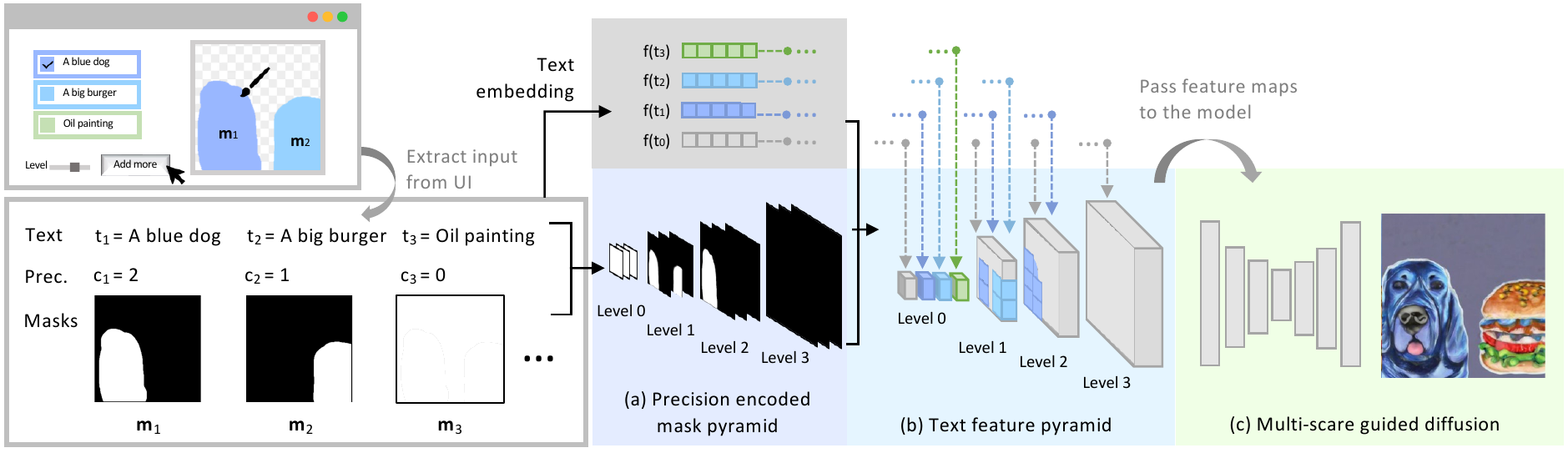}
\end{center}
\vspace{-10pt}
\caption{An overview of the proposed method. We provide an intuitive interface where users can easily define a layout using a semantic brush associated with a free-form text description and adjustable precision level. The masks, regional descriptions, and precision levels $\{(M_i, t_i, c_i)\}_{i=1}^{n}$ are jointly encoded into a text feature pyramid, and then translated into an image by a multi-scale guided diffusion model. }
\label{fig_model}
\end{figure*}

\section{Proposed Method}\label{sec_method}
Our method aims to generate images from a layout consisting of a set of semantic regions of varying precision levels. Formally, an input layout can be viewed as a list of tuples  $\{(M_i,t_i,c_i)\}_{i=1}^n$, where $M_i,t_i,c_i$ indicate the segmentation mask, text description, and the precision level of the $i$-th region. We let $\bm{M},\bm{t},\bm{c}$ denote the sets of masks, texts and precision levels of all regions, respectively. 
The precision level variable $c_i \in \mathbb{N}_{\le L}$ indicates how precisely the generated content should follow the mask $M_i$.  Smaller value of $c_i$ indicates a less precise control allowing more deviation from the mask $M_i$. $c_i=0$ indicates the coarsest level where the $i$-th mask will be ignored. When all $c_i=0$, the problem converts into T2I. 
Fig.~\ref{fig_model} shows an overview of the proposed method. 
To generate an image at resolution $2^L\times 2^L$, we first formulate a precision-encoded mask pyramid $\{\bm{M}^l\}_{l=0}^L$ which represents each mask at the given precision level  (Sec.~\ref{sec_mask_levels}). Then we combine the mask pyramid with the text descriptions $\bm{t}$ to form a text feature pyramid $\{ \bm{Z}^l\}_{l=0}^L$. It contains a $2^l\times 2^l$ text feature map at each level, which can be seen as an extension of the one-hot label map encoding in S2I (Sec.~\ref{sec_feat}). 
Finally, a multi-level guided diffusion model takes the feature pyramid as input to generate an image (Sec.~\ref{sec_model}). 

\subsection{Precision-Encoded Mask Pyramid}
\label{sec_mask_levels}
\begin{figure}[t]
% \vspace{-10pt}
\begin{center}
% \fbox{\rule{0pt}{1.5in} \rule{\linewidth}{0pt}}
\includegraphics[width=\linewidth]{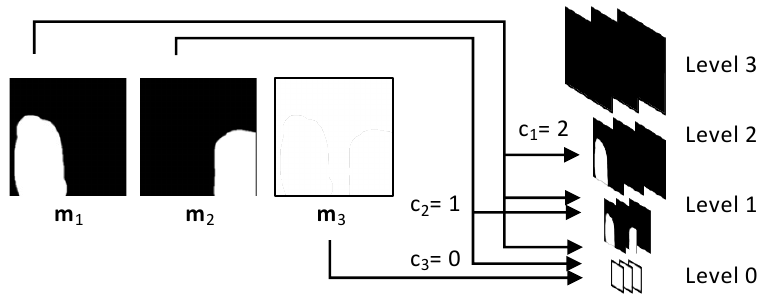}
\end{center}
\vspace{-10pt}
\caption{An example of the precision-encoded mask pyramid. 
The first level has two non-zero masks corresponding to $m_1,m_2$ as $c_1,c_2\ge 1$; the second level only has $m_1$ as $c_1=2,c_2<2$; the third level does not have any non-zero masks as all the precision levels of all masks are lower than $3$. 
}
\label{fig_featp}
\end{figure}
For a user-drawn coarse shape, it is challenging to model the exact precision, since the type and the amount of error vary across different users. To simplify the problem, we relate precision to resolutions and propose a precision-encoded mask pyramid to encode the shape and precision information simultaneously. Given a mask and a precision level, a high precision level corresponds to using all details of the masks at a high resolution, while a low precision level means we can only trust the mask at a low resolution. 

\begin{figure}[t]   
\begin{center}
% \fbox{\rule{0pt}{1.5in} \rule{\linewidth}{0pt}}
% \vspace{-5pt}
\includegraphics[width=\linewidth]{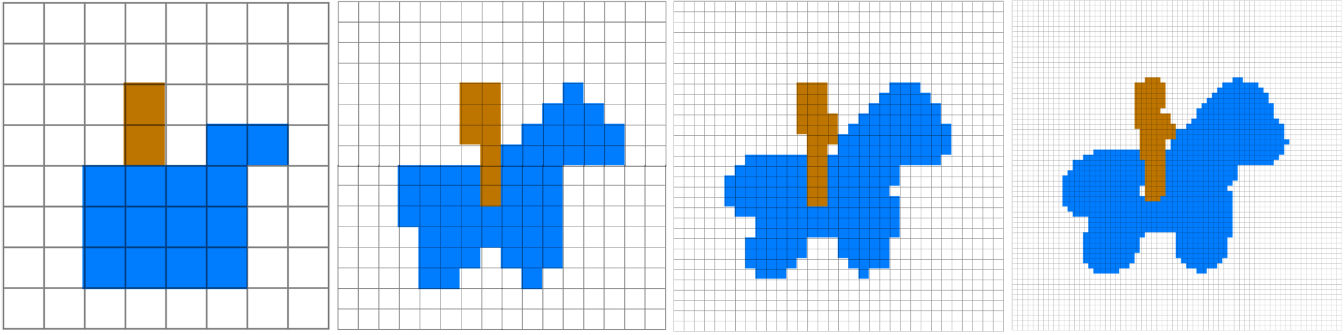}\\
    \scriptsize{\hfill {Level 3 ($8\times 8$)} \hfill\hfill {Level 4 ($16\times 16$)} \hfill\hfill {Level 5 ($32\times 32$)} \hfill\hfill {Level 6 ($64\times 64$)} \hfill\hfill}
\end{center}
\vspace{-10pt}
\caption{Illustration of the masks at different levels of a pyramid.
}
\label{fig_mask_levels}
\end{figure}

More specifically, given a set of masks $\bm{M}=[m_1,..,m_n]$ and the precision levels $\bm{c}=[c_1,...,c_n]$, we construct the mask pyramid by representing each mask $M_i$ at resolutions up to $2^{c_i}$. In other words, we resize each mask $M_i$ to multiple resolutions and drop out the masks at resolutions higher than $2^{c_i}$, as illustrated in Fig.~\ref{fig_featp}. 
Formally, the $l$-th layer of the mask pyramid, denoted as $\bm{M}^l$, is computed as follows, 
\setlength{\belowdisplayskip}{3pt} \setlength{\belowdisplayshortskip}{3pt}
\setlength{\abovedisplayskip}{3pt} \setlength{\abovedisplayshortskip}{3pt}
\begin{equation}
M_i^l = \mbox{resize}(M_i, 2^l) \cdot \mathbbm{1}_{c_i\ge l},
\end{equation}
where $\mbox{resize}(M_i,2^l)$ resizes a mask $M_i$ to $2^l\times 2^l$ resolution through image interpolation and binarization. 
The indicator function $\mathbbm{1}_{c_i\ge l}$ returns 1 when $c_i\ge l$ and $0$ otherwise. 
In our implementation, we set the range of precision levels to $[0,3,4,5,6]$. We observed that for a free-hand drawn layout, $8\times 8$ masks at the level $3$ are easy to draw while being fairly informative, whereas $64\times 64$ masks at the level $6$ usually capture enough details, as illustrated in Fig.~\ref{fig_mask_levels}.

\subsection{Text Feature Pyramid}\label{sec_feat}
The mask pyramid encodes the shape and the precision information. To generate an image, we also need semantic information. We introduce a text feature pyramid for this purpose. 
Each level of the text feature pyramid is a $2^l\times 2^l$ text feature map $\bm{Z}^l$ obtained by combining the masks $\bm{M}^l$ and the embeddings of the text $\bm{t}$. At the $0$-th level, the $1\times 1$ masks contain no shape information, so we simply concatenate the embeddings of all words into a sequence. At the levels where $l>0$, we spatially spread the embeddings of $\bm{t}$ over the corresponding masks to jointly represent the shape and semantic, as illustrated in Fig.~\ref{fig_model}. 

\noindent\textbf{Text feature maps. }
Here we describe in more detail how we construct the text feature map $\bm{Z}^l$ at level $l>0$. 
For simplicity, we drop the superscript $l$ and denote the masks at an arbitrary level as $\bm{M}$. Each element $M_{i,x,y}$ is a binary value $\in \{0,1\}$ and we allow overlaps or blank spaces,~\ie $n \ge \sum_i^n M_{i,x,y} \ge 0$. 
Given the initial masks $\bm{M}$, we introduce normalized masks  $\hat{\bm{M}}$, which are augmented from $\bm{M}$ by adding an extra mask $M_0$ to indicate the blank space and normalized by the number of shapes at each location:
\begin{equation}
\begin{split}
\hat{M}_{i,x,y} &= M_{i,x,y} / \sum_{i=0}^n M_{i,x,y},\\
\mbox{where}~M_{0,x,y}&=\mathbbm{1}_{\sum_{i=1}^n M_{i,x,y}=0}.
\end{split}
\end{equation}
In the normalized masks $\hat{\bm{M}}$, each element $\hat{M}_{i,x,y}$ is a continuous value $\in [0,1]$ and we have $\sum_{i=0}^n \hat{M}_{i,x,y}=1$. We compute each element $\bm{Z}_{x,y}$ at location $x,y$ as follows,
\begin{equation}
\label{eq_vec_to_text_map}
\begin{split}
\bm{Z}_{x,y} &= \sum_{i=0}^n \bm{f}(t_{i}) \cdot \hat{M}_{i,x,y},
\end{split}
\end{equation}
where $\bm{f}(t_i)$ is an embedding of the text $t_i$. We set $t_0$ to be a null token $\oslash$ to represent unspecified areas,~\ie the blank space indicated by $M_0$. 

The text feature map representation has several advantages. First, it is of the same dimension regardless of the number of masks, and therefore compatible with most deep ConvNet architectures. Second, each element of a text feature map is a convex combination of $n$ text embeddings in the learned embedding space. With a powerful language model, we can achieve a good generalization capability for unseen combinations of concepts. In a text feature map, any overlapping area contains an interpolation of multiple embeddings. Accordingly, users can get creative results derived from hybrid concepts by drawing overlapping shapes. 

\noindent\textbf{Data acquisition. }
During inference, the segmentation masks and text descriptions are both provided by users. During training, we can generate them automatically using text-based object  detection~\cite{li2021grounded} and segmentation~\cite{he2017mask}. We set the regions where no objects are detected as blank space and assign a null token $\oslash$ to these regions. 
We use CLIP~\cite{radford2021learning} text model to encode the text descriptions. 
In addition, the $0$-th level feature maps can be obtained directly from the image caption embeddings. 
More details regarding the training data generation step can be found in the supplementary material. 

\noindent \textbf{Relation to one-hot label maps.} In S2I, a layout containing at most $C$ classes can be encoded into a $C$-channel one-hot label map. It can be seen as a special case of text feature maps when the masks are dense and non-overlapping, and the embedding model $\bm{f}$ is a one-hot encoding function. Specifically, let $\bm{M}$ be the segmentation masks, $\bm{t}$ the class labels where $t_i=1,2,...,C$, $f_c(t_i)=\mathbbm{1}_{t_i=c}$. Since $\bm{M}$ are dense and non-overlapping, the normalized masks $\hat{M}_i$ will be the same as $M_i$ for $i>1$ and $\hat{M}_0=0$. Eqn.~\ref{eq_vec_to_text_map} then becomes
\begin{equation}
Z_{c,x,y} = \sum_{i=1}^n \mathbbm{1}_{t_i=c} \cdot M_{i,x,y} = \sum_{i | t_i=c}M_{i,x,y}.
\end{equation}
Therefore, the $c$-th channel of $\bm{Z}$ is a mask covering all pixels of class $c$,~\ie a one-hot label map. 
Through this re-formulation, we can see that the one-hot  representation has limited capability as $\bm{f}(t_i)$ is restricted to be binary. The text feature map representation uses a learned language model as $\bm{f}$ to encode the more informative open-domain layouts. 

\subsection{Multi-Scale Guided Diffusion }
\label{sec_model}
\begin{figure}[t]
\begin{center}
% \fbox{\rule{0pt}{1.5in} \rule{\linewidth}{0pt}}
\vspace{-10pt}
\includegraphics[width=\linewidth]{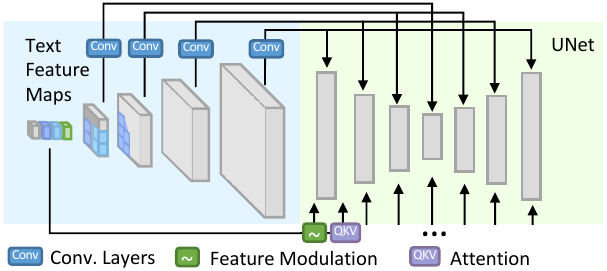}
\end{center}
\vspace{-10pt}
\caption{Architecture of the multi-scale guided diffusion model. 
}
\label{fig_unet}
\end{figure}
We use a diffusion model to generate images from a text feature pyramid. We use $\bm{\epsilon}$-prediction and the simplified training objective following~\cite{ho2020denoising}:
\begin{equation}
L = \mathbb{E}_{\bm{\epsilon}\sim \mathcal{N}(0,1),t\sim[1,T]} \left [ \| \bm{\epsilon} - \bm{\epsilon_\theta} (\bm{x}_t,\bm{z},t)\|_2^2 \right ],
\end{equation}
where $T$ is the number of diffusion steps, $\bm{x}_t$ is a noisy version of the ground-truth image $\bm{x}$ at the $t$-th diffusion step, $\bm{z}$ is the conditional signal, which in our case is the text feature pyramid $\{\bm{Z}^l\}_{l=0}^L$, and $\bm{\epsilon_\theta}$ represents the network parameterized by $\bm{\theta}$. 
We modify the UNet architecture~\cite{dhariwal2021diffusion} by adding the convolutional layers to combine each text feature map with the UNet feature maps of the corresponding resolution. 
The $0$-th level text feature maps $\bm{Z}^0$ are passed through all blocks using cross-modal attention~\cite{saharia2022photorealistic} and channel-wise feature modulation. 
Fig.~\ref{fig_unet} shows the overall architecture of the modified UNet. 

Classifier-free guidance~\cite{ho2021classifier} has shown to be an effective technique to improve the performance of conditional diffusion models and is widely applied to diffusion models conditioned on text and class tags~\cite{dhariwal2021diffusion,saharia2022photorealistic,nichol2021glide,rombach2022high}. To benefit from this technique, we introduce multi-scale spatial guidance for our feature map conditioned diffusion model. As described in Sec.~\ref{sec_feat}, in text feature maps from level $1$ to $L$, the unspecified regions are the embedding of a null token $\bm{f}(\oslash)$. We further apply dropout with probability $0.1$ to $\bm{Z}^0$ by setting $\bm{Z}^0=\bm{f}(\oslash)$.  
During inference, we estimate two diffusion scores, conditioning on the given text feature pyramid and an empty feature pyramid of repeating $\bm{f}(\oslash)$, and then perform sampling using their linear combination.

\section{Experiments}
\label{sec_exp}
In this section, we present the experimental results on image synthesis from any-level semantic layouts. 
We evaluate the proposed method on user-drawn coarse layouts and automatically simulated layout data. We also present results on T2I and S2I, and compare our method with state-of-the-art methods. 

\subsection{Implementation Details}
We train the multi-scale guided diffusion model to generate $64\times 64$ images from a layout of precision up to $\log_2 64$. To generate images at higher resolution, we experiment with two approaches: (1) by training another diffusion model to upsample $64\rightarrow 256$ following~\cite{dhariwal2021diffusion}; (2) by generating the latent map at $64\times 64$ resolution then decoding it into a $512\times 512$ image  following~\cite{rombach2022high}. 
Both approaches are effective, demonstrating that the proposed method can generalize well to both pixel-space and latent-space diffusion strategies. We use smaller pixel-space diffusion models for ablation studies and report the main results based on a larger latent-space diffusion model. 
We encode the text descriptions using the pretrained CLIP ViT-L14 language model~\cite{radford2021learning}. To avoid repeated forward passes, we feed the union of all regional descriptions $\cup_i t_i$ as a sentence into the language model. 
Then, we average the hidden states corresponding to each word in $t_i$ to use as its embedding $\bm{f}(t_i)$. 

\noindent\textbf{Training details. } We use the aesthetic subset of Laion2B-en dataset~\cite{schuhmann2022laion} for training. 
We generate full text feature pyramids for 5M randomly selected samples and construct only the $0$-th level feature maps for the remaining samples. 
For pixel space diffusion, we train a base model of 300M parameters and a super-resolution model of 300M parameters; for latent-space diffusion, we train a base model of 900M parameters and use the pretrained deocder from~\cite{rombach2022high}. 
We use batch size $2048$ for $64\times 64$ models and batch size $960$ for the super-resolution model. 

\begin{figure}[h]
\begin{center}
\vspace{-10pt}
% \fbox{\rule{0pt}{1.5in} \rule{\linewidth}{0pt}}
\includegraphics[width=\linewidth]{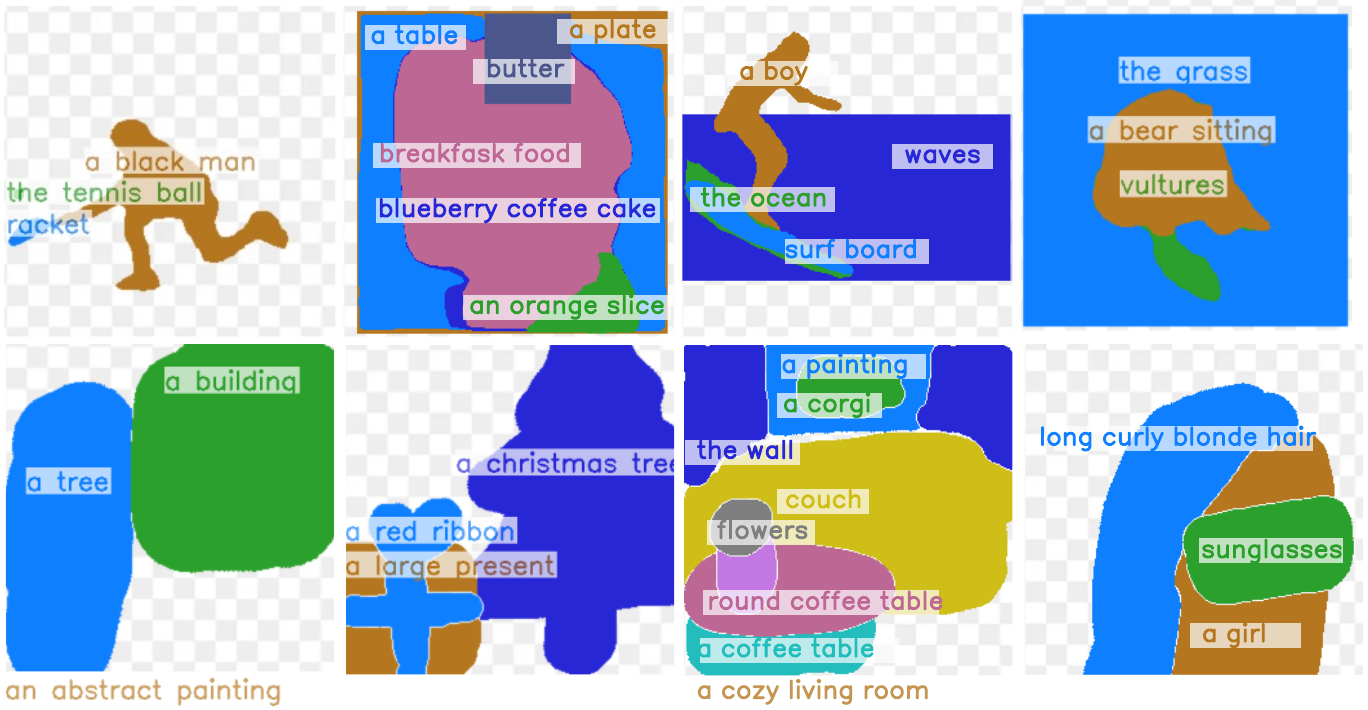}
\end{center}
\vspace{-10pt}
\caption{Top: Pseudo layouts from OpenLayout-COCO dataset. Bottom: real-world user-drawn layouts from OpenLayout dataset. 
}
\label{fig_benchmark}
\end{figure}

\noindent\textbf{Evaluation benchmark. }
We use the COCO~\cite{lin2014microsoft} validation set for evaluation in the T2I setting and the COCO-stuff~\cite{caesar2018coco} validation set for S2I. 
For evaluation in open-domain layout-to-image generation, we construct a new test set OpenLayout containing 260 user-drawn layouts of coarse shapes. 
The layouts are annotated by 10 users based on text prompts randomly sampled from PartiPrompts~\cite{yu2022scaling}. 
To analyze the effect of precision level control, we also evaluate using the pseudo layouts consisting of accurate shapes, which are extracted from the 5,000 images with caption annotations of COCO validation set (OpenLayout-COCO). 
Fig.~\ref{fig_benchmark} shows some layouts samples from OpenLayout and Openlayout-COCO. 

\subsection{Open-Domain Layout-to-Image Generation}
\label{sec_exp_ov2im}
\begin{figure*}[t]
\begin{center}
\vspace{-10pt}
% \fbox{\rule{0pt}{3in} \rule{\linewidth}{0pt}}
\includegraphics[width=\linewidth]{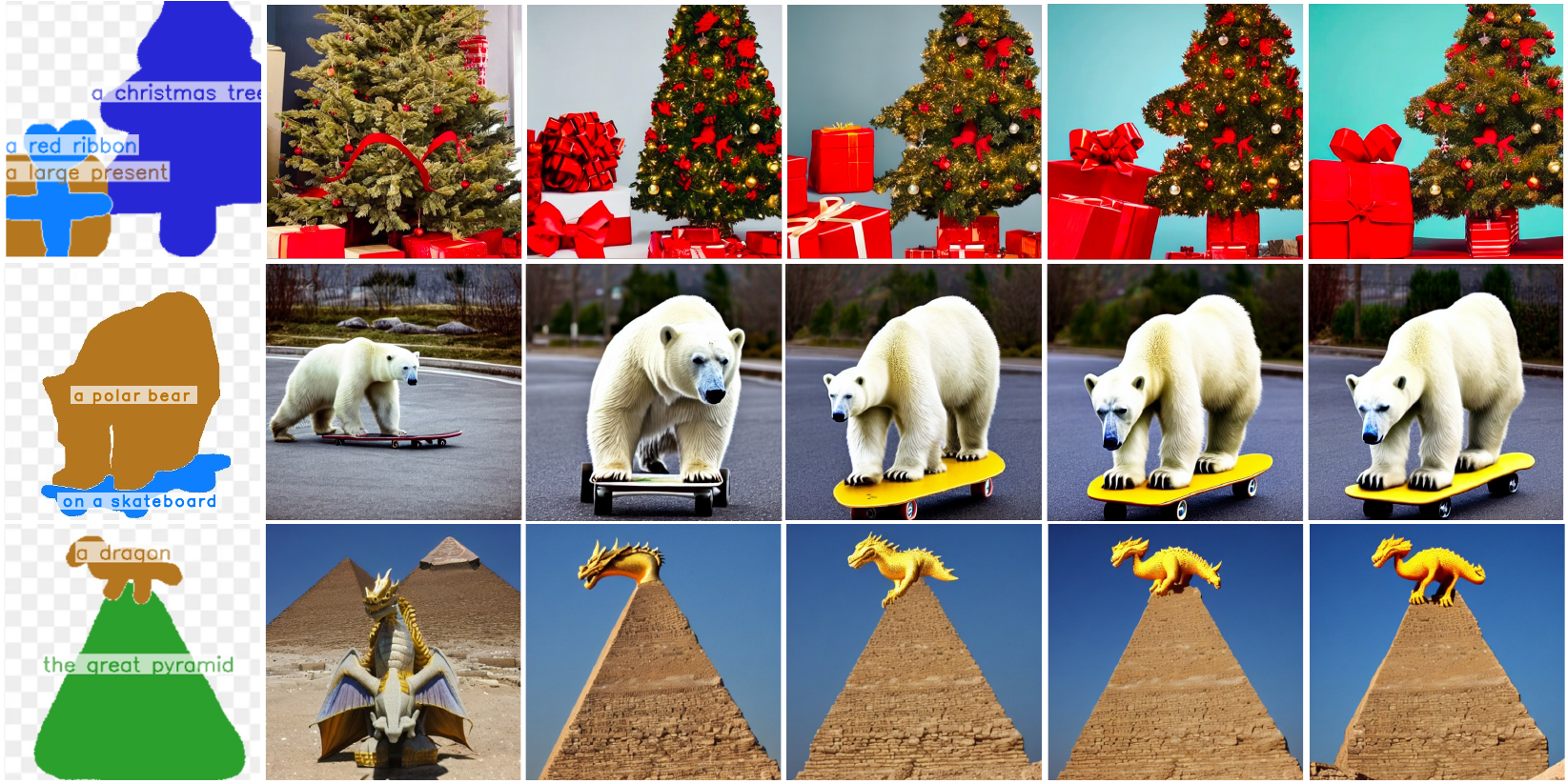}\\
\scriptsize{ \hfill {Input} \hfill\hfill {Level 0} \hfill\hfill {Level 3} \hfill\hfill {Level 4} \hfill\hfill {Level 5} \hfill\hfill {Level 6} \hfill\hfill}
\end{center}
\vspace{-10pt}
\caption{Results with different precision levels. For each input layout, we sample the images starting from same noise, so the images at different precision levels can have similar styles. 
}
\label{result_layout}
\end{figure*}
\begin{figure}[t]
\begin{center}
\vspace{-10pt}
% \fbox{\rule{0pt}{3in} \rule{\linewidth}{0pt}}
\includegraphics[width=\linewidth]{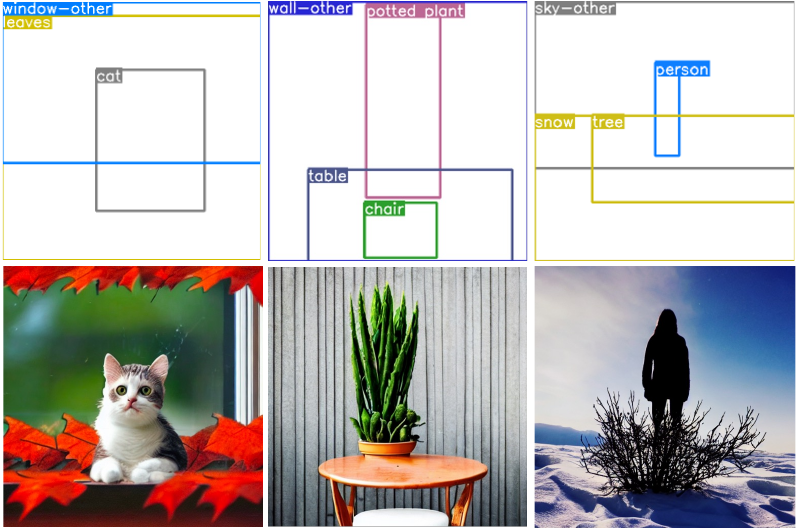}
\end{center}
\vspace{-10pt}
\caption{Results with bounding box layout. Precision level $=3$. 
}
\label{result_bo_layout}
\end{figure}
Here we discuss the results on the OpenLayout and OpenLayout-COCO datasets. For each sample, we run the model a varying precision level $c$ from $0$ to $6$ to verify the effectiveness of precision control. Fig.~\ref{result_layout} shows the results with different precision levels. For simplicity, we use the same precision level for all regions in an input layout,~\ie $\forall i, c_i=c$. We can see that as the precision level increases, the generated images follows the layout more closely. When $c=0$, the image compositions are not related to the layouts. For $c=3,4$ the generated images roughly resemble the shape and location specified in the layout. For $c=5,6$, the generated object contours matches the layout more closely.
At the lowest precision level, our method can handle very rough layouts,~\eg, bounding box layouts, as shown in Fig.~\ref{result_bo_layout}, despite not being trained on bounding box data. 
\begin{table}[t]
\caption{\small Quantitative results on the OpenLayout and OpenLayout-COCO datasets with different precision levels. SS Score: spatial similarity scores.}
\begin{center}
\label{table_layout}
\vspace{-10pt}
\resizebox{\columnwidth}{!}{
\begin{tabular}{ccc|ccc}
\hline
 &\multicolumn{2}{c|}{OpenLayout}&\multicolumn{3}{c}{OpenLayout-COCO}\\
\hline
Prec. & SS Score $\uparrow$ &CLIP Score $\uparrow$ & SS Score $\uparrow$ & CLIP Score $\uparrow$ & FID $\downarrow$ \\
\hline
0 & .436& \textbf{.274} & .572& \textbf{.260}& \textbf{25.3}\\
3 & .533& .268 & .685& .256& 26.0\\
4 & .543& .266 & .716& .256& 27.0\\
5 & .558& .267 & .729& .256& 28.4\\
6 & \textbf{.558}& .261 & \textbf{.736} & .255& 30.6\\
\hline
\end{tabular}
}
\vspace{-10pt}
\end{center}
\end{table}

For quantitative evaluation, we compute the CLIP score using the original captions to measure global semantic alignment. To measure spatial alignment, we define a spatial similarity score (SS score), which is the cosine similarity between the text feature maps of the input layout and the layout reconstructed from the generated image. For OpenLayout-COCO, as the ground-truth images are available, we also compute the FID to measure the visual quality. 

Table~\ref{table_layout} reports the quantitative evaluation results on images generated with different precision levels. The second and fourth columns show that using a higher precision level generally leads to a higher spatial similarity, which further demonstrates the effectiveness of precision control. On OpenLayout-COCO, the spatial similarity consistently increases as the precision level becomes higher. 
Whereas, on OpenLayout, it stays the same at higher levels. This is due to the difference in the inherent layout precision exhibited by the two datasets; compared to OpenLayout-COCO, the layouts in OpenLayout are coarser. For a coarse layout, the generated images already match it well at low precision levels and the room for improvement is limited. 

From the last column of Table~\ref{table_layout}, we can see that the results with a lower precision level have a smaller FID, which indicates a similar distribution with respect to the ground-truth images. This is likely because a lower precision level enforces smaller constraints on the generation process, and therefore the generated images can better capture the real image distribution.  
Similarly, a lower precision level also yields a slightly better CLIP score due to the smaller spatial constraints, as shown in the third and fifth columns of Table~\ref{table_layout}. 

\subsection{Text-to-Image Generation}
\begin{figure*}[t]
\begin{center}
%\fbox{\rule{0pt}{1.5in} \rule{\linewidth}{0pt}}
\vspace{-10pt}
\includegraphics[width=\linewidth]{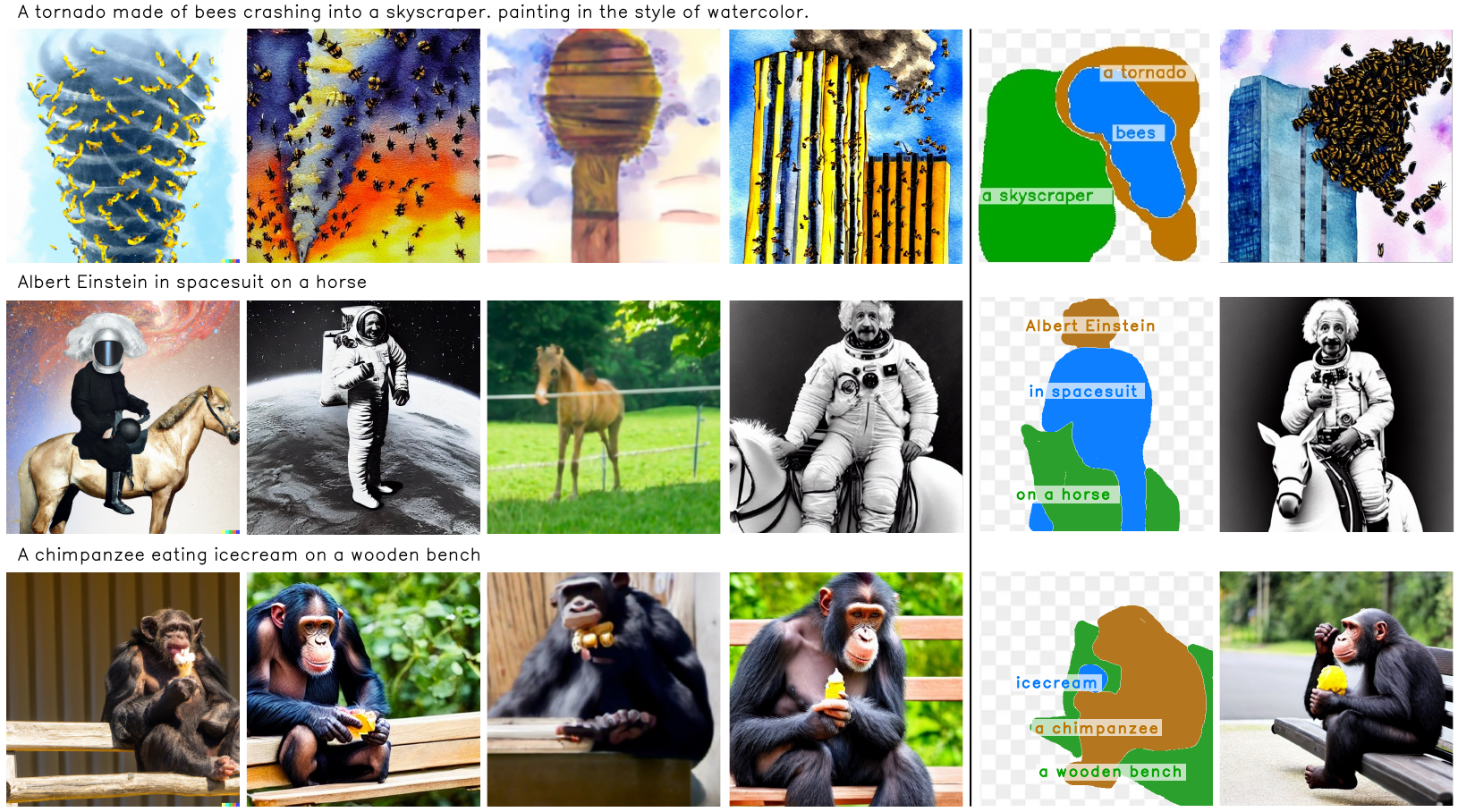}\\
\scriptsize{ \hfill {DALLE2} \hfill\hfill {\textcolor{white}{1}\textcolor{white}{1}SD\textcolor{white}{1}\textcolor{white}{1}} \hfill\hfill {\textcolor{white}{1}GLIDE} \hfill\hfill {\textcolor{white}{1}Ours} \hfill\hfill {Layout} \hfill\hfill {\textcolor{white}{1}Ours$^*$\textcolor{white}{1}} \hfill\hfill}
\end{center}
\vspace{-10pt}
\caption{Text-to-image generation results. Ours: our results by using text as a $0$-th level layout. Ours$^*$: our results with layout control. 
}
\label{result_text2im_layout}
\end{figure*}
As mentioned in Sec.~\ref{sec_method}, our method can be applied for text-to-image generation by using texts as $0$-th layouts. 
The first four columns of Fig.~\ref{result_text2im_layout} show the images generated from text prompts by our method and state-of-the-art methods. We can see that the proposed method can generate visually peasant images with reasonable layouts from only text input.  Table~\ref{table_text2im} reports the FID evaluated on the COCO validation set. The classifier-free guidance scale is set to $3$. Following~\cite{nichol2021glide}, we sample 30K images using randomly selected text prompts and compute FID against the entire validation set. 
It can be seen that the proposed method compares favorably against state-of-the-art text-to-image generation models. 
\begin{table}[t]
\caption{\small Quantitative comparison with T2I methods on COCO.  }
\vspace{-5pt}
\begin{center}
\label{table_text2im}
% \resizebox{\columnwidth}{!}{
% }
\begin{tabular}{ccc}
\hline
Method & FID $\downarrow$ & zero-shot FID$\downarrow$\\
\hline
% PDM~\cite{} &5.34 & 13.28\\
SSA-GAN~\cite{liao2022text} & 19.37 & - \\
VQ-Diffusion~\cite{gu2022vector} & 13.86 & - \\
DF-GAN~\cite{tao2022df} & 19.32 &  - \\
% LAFITE~\cite{} &8.12 & 26.94\\
\hline
% PDM~\cite{} && 13.28\\
% LAFITE~\cite{} && 26.94 \\
% LDM~\cite{rombach2022high} &  -& 12.63 & x\\
% MakeAScene~\cite{gafni2022make} & & 11.84 & x\\
GLIDE~\cite{nichol2021glide} &  -& 12.89 \\
SD~\cite{rombach2022high,stablediffusion} &- & 9.89 \\
DALLE-2~\cite{ramesh2022hierarchical} &- & 10.87\\
Ours & \textbf{8.55} & \textbf{9.47} \\ 
% Ours &- & 9.47 \\ 
\hline
\end{tabular}
\vspace{-10pt}
\end{center}
\end{table}

By combining with a named entity recognition (NER) model, we can apply the proposed method to layout controllable text-to-image generation. More specifically, given an input sentence, we parse the noun phrases using NER to generate regional text descriptions, and users can arbitrarily draw the shapes for those noun phrases.  The last two columns of Fig.~\ref{result_text2im_layout} show the text-to-image generation results with layout control. We can see that the generated images match the text well and follow the provided layouts. 

\subsection{Segmentation-to-Image Generation}
\begin{figure*}[t]
\begin{center}
% \fbox{\rule{0pt}{1.5in} \rule{\linewidth}{0pt}}
\vspace{-5pt}
\includegraphics[width=\linewidth]{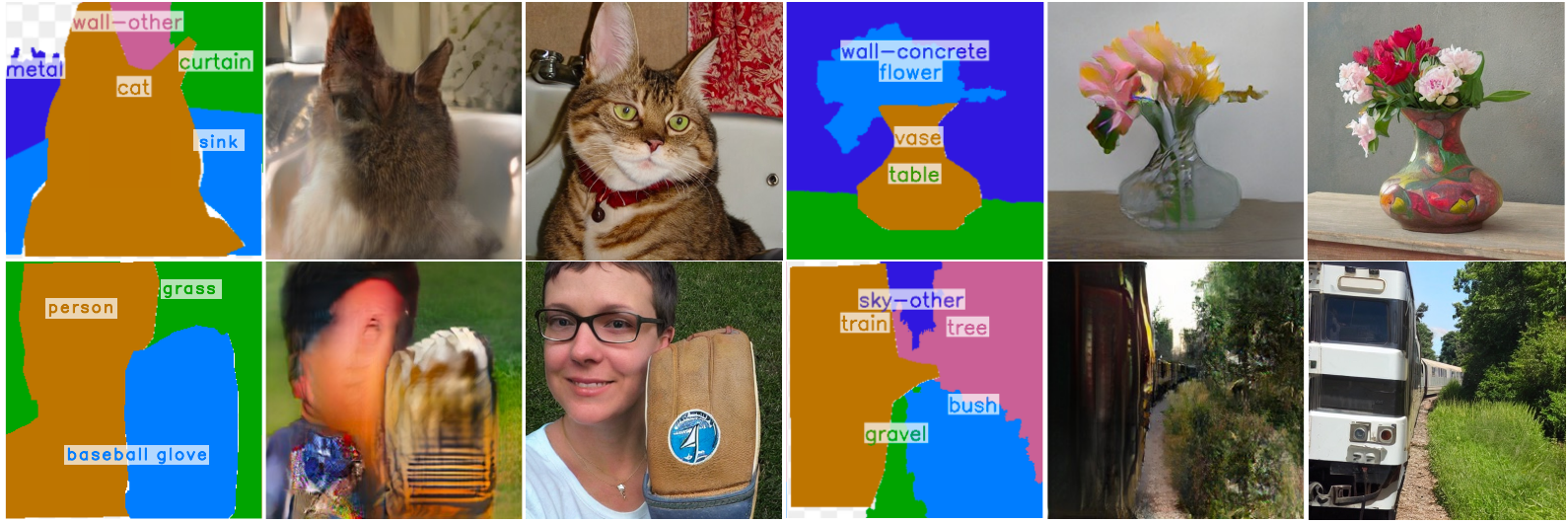}\\
\scriptsize{ \hfill {Input} \hfill\hfill {SPADE} \hfill\hfill {\textcolor{white}{1}Ours} \hfill\hfill {Input} \hfill\hfill {SPADE} \hfill\hfill {Ours\textcolor{white}{1}} \hfill\hfill}
\end{center}
\vspace{-10pt}
\caption{Visual comparison with SPADE~\cite{park2019semantic} for segmentation-to-image generation. 
}
\label{result_seg2im}
\end{figure*}

Our model can also generate images from a dense segmentation map of closed-set labels,~\ie following the original S2I setting. For this application, we treat the class labels as text descriptions and use the highest precision level $c=5$ for all masks. Fig.~\ref{result_seg2im} shows example images generated by our method and SPADE~\cite{park2019semantic}. The images generated by our model are of significantly better visual quality. 
Table.~\ref{table_seg2im} reports the quantitative comparison results on the COCO-stuff validation set with the state-of-the-art S2I methods~\cite{park2019semantic,tan2021efficient,lv2022semantic}. Since the highest resolution of the input layout to our model is $64$, we evaluate the S2I results at $64\times 64$ resolution. 
From the third and fifth columns of Table.~\ref{table_seg2im} we can see that despite not being designed or trained for this task, the proposed method can achieve lower FID and comparable mIOU. After being fine-tuned on the COCO-stuff training set, our model outperforms previous approaches in terms of both FID and mIOU. 
\begin{table}[h]
\caption{\small Quantitative comparison with S2I methods on COCO-stuff. }
\vspace{-10pt}
\begin{center}
\label{table_seg2im}
\resizebox{\columnwidth}{!}{
\begin{tabular}{cccccc}
\hline
Method & FID $\downarrow$ & zero-shot FID $\downarrow$ & mIOU$\uparrow$ & zero-shot mIOU$\uparrow$\\
\hline
SPADE~\cite{park2019semantic} & 50.91 &- &  17.49 &-\\
SAFM~\cite{lv2022semantic} & 62.28&- &12.28 & - \\
CLADE~\cite{tan2021efficient} & 55.30 & - &17.21 & - \\
\hline
Ours &\textbf{17.20} & 20.17&\textbf{23.01}& 17.15\\
\hline
\end{tabular}
}
\vspace{-10pt}
\end{center}
\end{table}

\subsection{Ablation Studies}
We compare the proposed any-level method with two fixed-level baseline models: a text-to-image generation model and a fixed-level layout-to-image generation model. The baseline models are of the same architecture as the any-level models. When training the baseline models, we only use layouts of fixed levels,~\ie the $0$-th level for the text-to-image baseline and the $4$-th level for the fixed-level segmentation-to-image baseline. Table~\ref{table_abl} compares the evaluation results of the any-level model and fixed-level baselines at the corresponding levels. The any-level model achieves better results than the baseline models, which further demonstrates the advantage of the unified framework. 

\begin{table}[t]
\vspace{-10pt}
\caption{\small Ablation study results on OpenLayout-COCO.}
\vspace{-5pt}
\begin{center}
\label{table_abl}
\resizebox{\columnwidth}{!}{
\begin{tabular}{ccccc}
\hline
Prec. Level & Model & FID $\downarrow$ & CLIP Score $\uparrow$ & SS Score $\uparrow$\\
\hline
0& Text-to-Image Baseline &32.83 &.2473 & .5613 \\
0 &Any-Level Model & \textbf{32.09}& \textbf{.2496}& \textbf{.5640}\\
\hline
4& Single-Level Baseline & 36.03& .2381& .5729\\
4&Any-Level Model & \textbf{33.70}& \textbf{.2475}& \textbf{.6978}\\
% Ours-Latent &4 & & & \\
% Ours$^\dagger$-70M & & & \\
% Ours$^\dagger$-45M & & & \\
% Ours$^\dagger$-5M & & & \\
% Ours-5M & & & \\
\hline
\end{tabular}
}
\vspace{-10pt}
\end{center}
\end{table}

\section{Conclusion}
This paper presents a new conditional image synthesis framework to generate images from any-level open-domain semantic layouts. The input level ranges from pure text to a 2D semantic canvas with precise shapes. Several novel techniques are introduced, including a pipeline for collecting training data; the representations to jointly encode precision level, semantics, and geometry information; and a multi-scale guided diffusion model to synthesize images. A test dataset containing user-drawn layouts is collected to evaluate the proposed method. Experimental results demonstrate the advantage of the unified framework. The proposed method can generate high-quality images following the layout at specified precision levels, and compares favorably against the state-of-the-art methods on public benchmarks.

\cleardoublepage
\setcounter{page}{1}
\setcounter{section}{0}
\setcounter{figure}{0}
\setcounter{table}{0}
\renewcommand\thesection{\Alph{section}} 
\renewcommand\thefigure{\Alph{figure}} 
\renewcommand\thetable{\Alph{table}} 

\onecolumn
% \title{SceneComposer: Any-Level Semantic Image Synthesis\\Supplementary Material}
% \maketitle

\section*{\centering Supplementary Material}
\section{Concepts Interpolation }
\begin{figure*}[h]
\begin{center}
%\fbox{\rule{0pt}{1.5in} \rule{\linewidth}{0pt}}
% \vspace{-10pt}
\includegraphics[width=\linewidth]{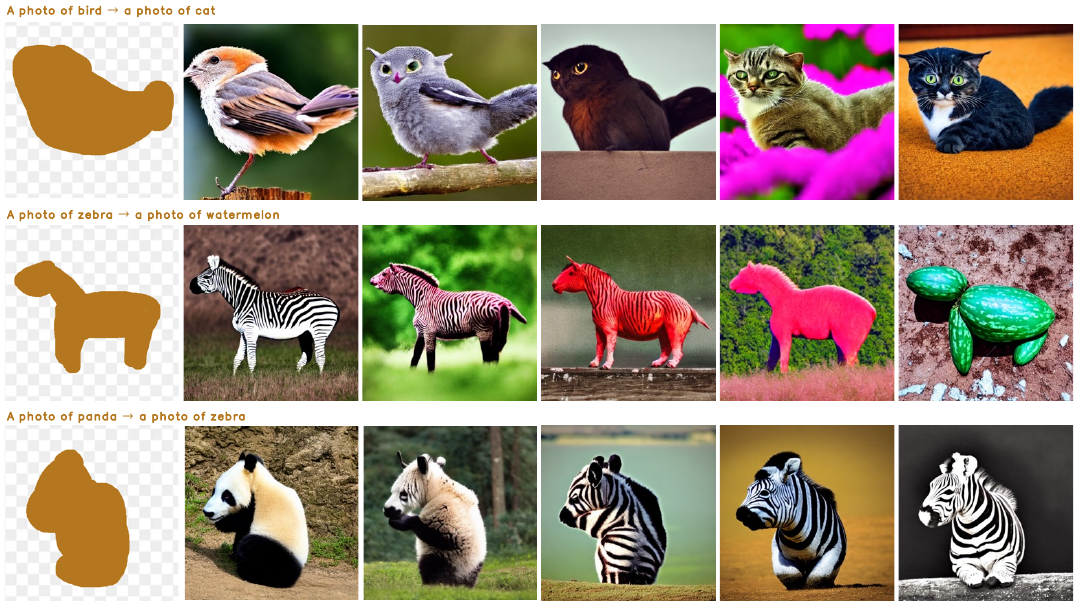}\\
% \scriptsize{\hfill\hfill\hfill {(a)} \hfill\hfill {(b)} \hfill\hfill {(c)} \hfill\hfill\hfill\hfill}
\end{center}
\vspace{-10pt}
\caption{Images generated from mixtures of two concepts. The weights for the second concepts gradually increase from left to right. 
}
% \vspace{-30pt}
\label{fig_mix_concept}
\end{figure*}

With the proposed text feature map representation, we can create mixtures of concepts using overlapping shapes with different text descriptions. We can change the importance of different concepts in the generated image by adjusting their weights.
Fig.~\ref{fig_mix_concept} shows example images generated from mixtures of concepts. We can see the gradual transition from the first concept to the second as the weights change. 

\section{Image Editing/Inpainting Results } 
\begin{figure*}[h]
\begin{center}
%\fbox{\rule{0pt}{1.5in} \rule{\linewidth}{0pt}}
% \vspace{-10pt}
\includegraphics[width=\linewidth]{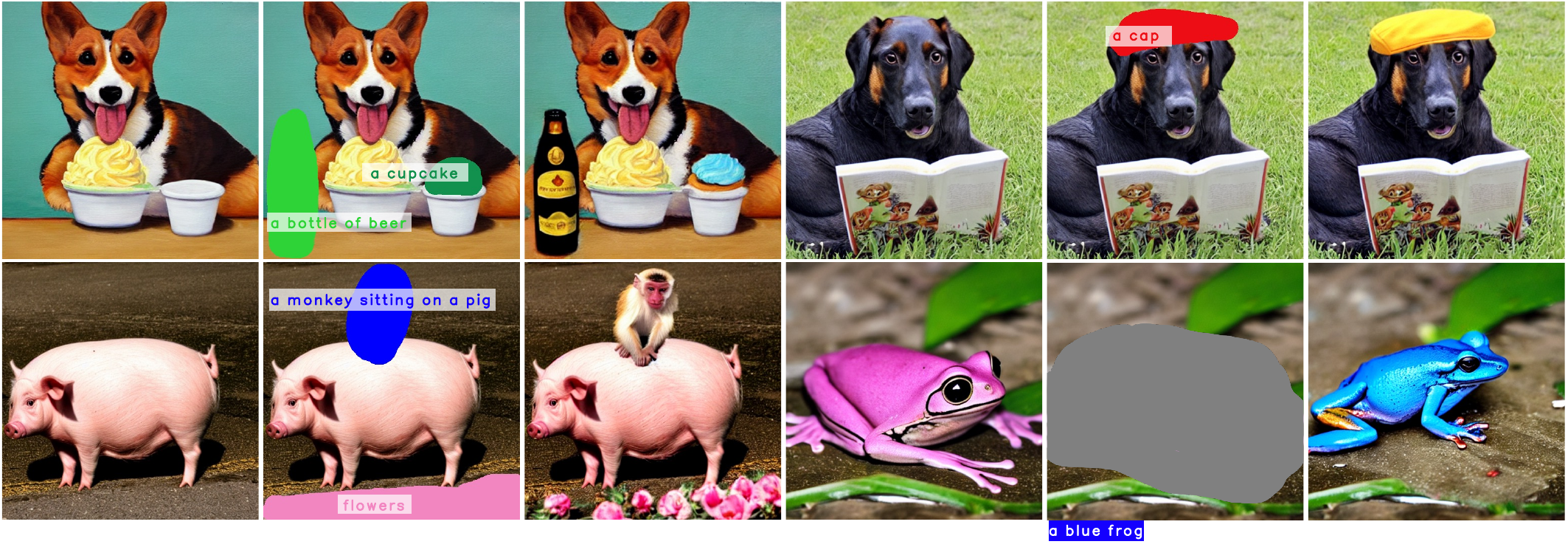}\\
\scriptsize{\hfill {Image} \hfill\hfill {Layout} \hfill\hfill {Result} \hfill\hfill {Image} \hfill\hfill {Layout} \hfill\hfill {Result} \hfill\hfill}
\end{center}
\vspace{-10pt}
\caption{Image inpainting/editing results. 
}
% \vspace{-30pt}
\label{fig_inpaint_result}
\end{figure*}

Given an image and a mask, the proposed method can be applied for image editing/inpainting by sampling the unmasked regions using the information in the original image, as in~\cite{lugmayr2022repaint}. More specifically, at each diffusion step, the masked regions are filled with the model output and the unmasked regions are from the noise corrupted original image. 
Fig.~\ref{fig_inpaint_result} shows the editing/inpainting results. 
For this application, we can also specify the layout and the precision levels. When the precision level $>0$, our method provides an intuitive image editing tool that allows users to edit by painting with a semantic brush. 
When the precision level is set to $0$, our method can inpaint a masked region (the frog example in Fig.~\ref{fig_inpaint_result}). 

\section{Interactive Image Synthesis}
\begin{figure*}[h]
\begin{center}
%\fbox{\rule{0pt}{1.5in} \rule{\linewidth}{0pt}}
% \vspace{-10pt}
\includegraphics[width=.6\linewidth]{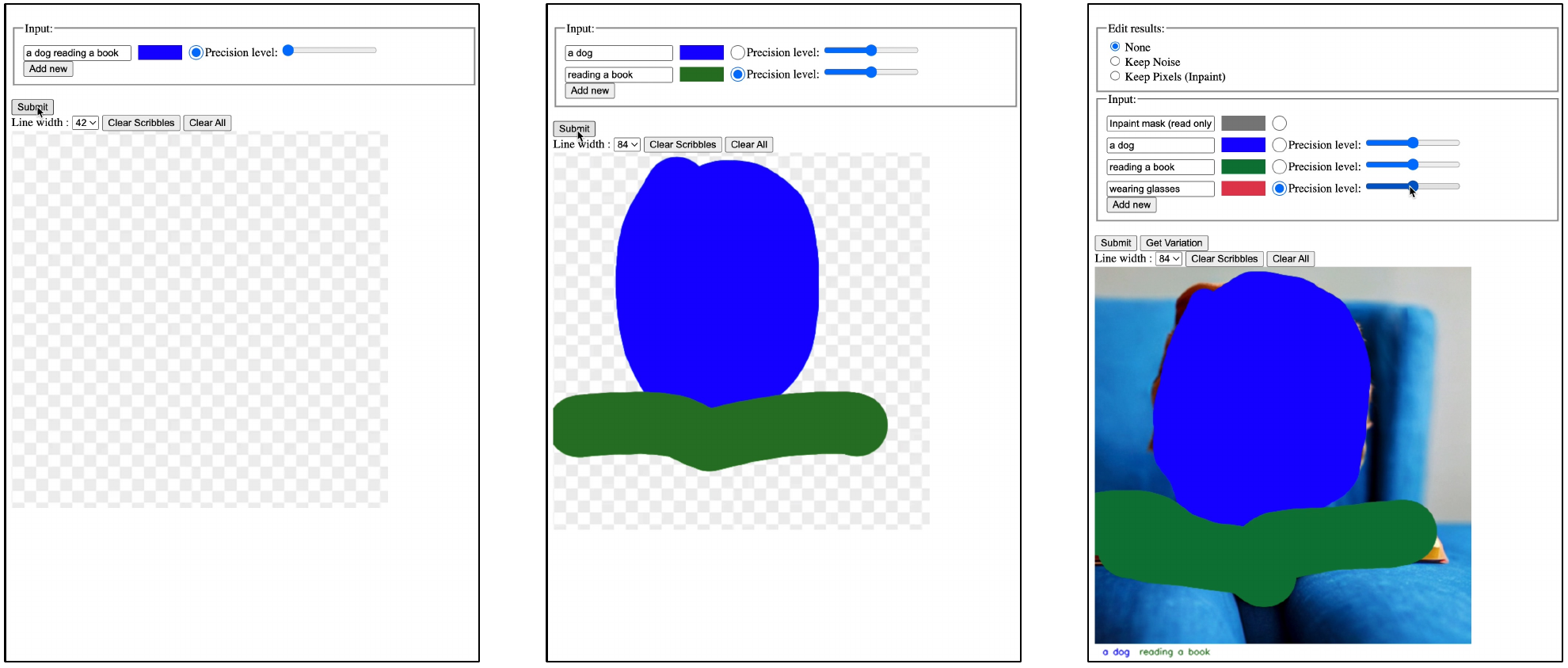}\\
\scriptsize{\hfill\hfill\hfill {(a)} \hfill\hfill {(b)} \hfill\hfill {(c)} \hfill\hfill\hfill\hfill}
\end{center}
\vspace{-10pt}
\caption{The user interface for interactive image synthesis. 
}
% \vspace{-30pt}
\label{fig_interface}
\end{figure*}

We build an interactive image synthesis application using the proposed framework.  Fig.~\ref{fig_interface} shows the user interface at different statuses. At the initial status (Fig.~\ref{fig_interface}~(a)), users define semantic concepts with free-from texts and may adjust the precision levels optionally. Then they can draw a layout on the blank canvas, or choose the lowest precision level and directly click submit (Fig.~\ref{fig_interface}~(b)). The generated images will be depicted on the canvas to allow further refinements/editing (Fig.~\ref{fig_interface}~(b)). At this stage, the users can choose to keep the noise from which the images are sampled to get an image of a similar style, or choose the inpainting mode to edit the results. More details regarding the interactive image synthesis application can be found in the demo video. 

\section{Results with Mixed Precision Levels} 
\begin{figure*}[h]
\begin{center}
%\fbox{\rule{0pt}{1.5in} \rule{\linewidth}{0pt}}
% \vspace{-10pt}
\includegraphics[width=\linewidth]{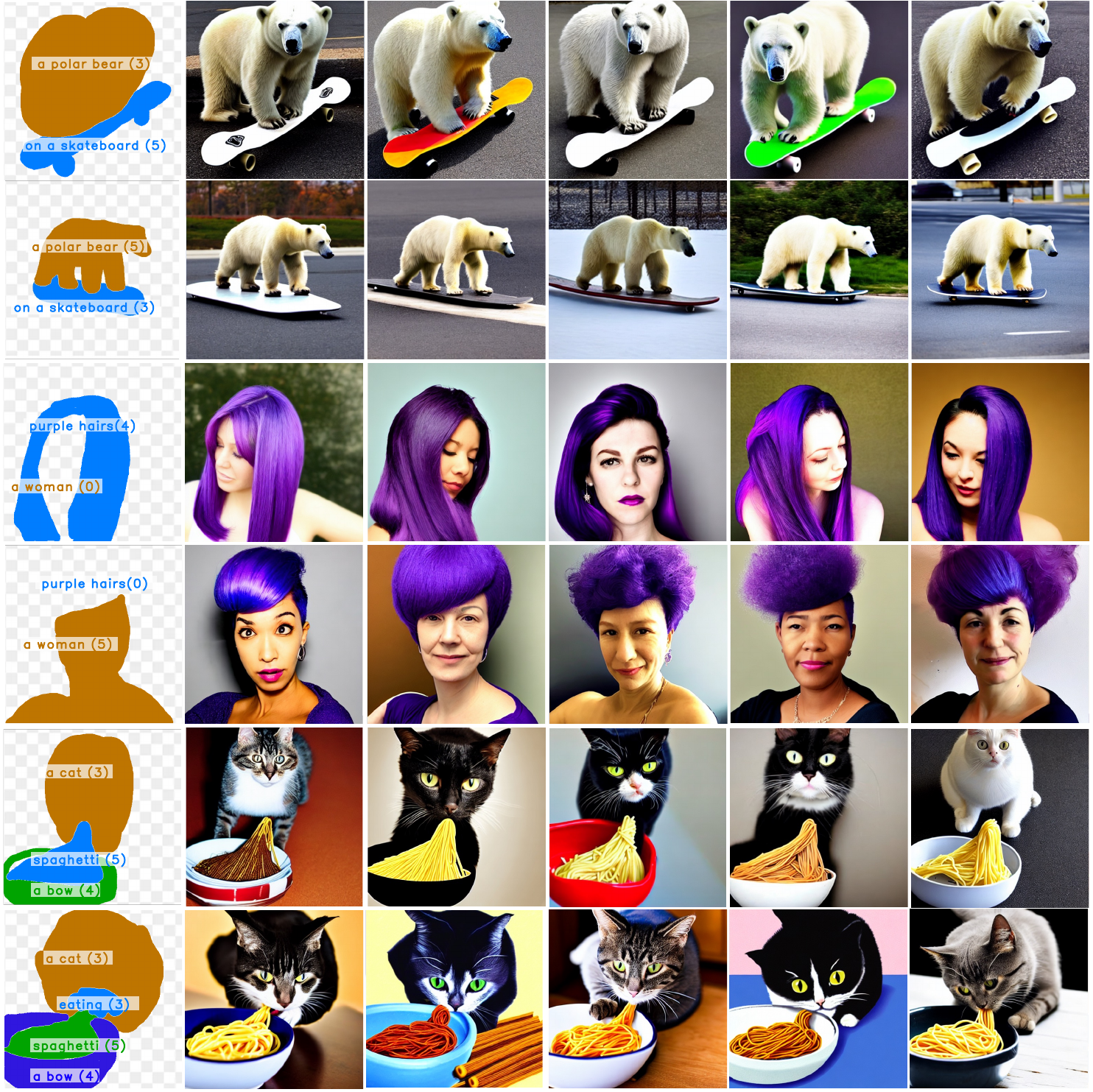}\\
% \scriptsize{ \hfill {Input} \hfill\hfill {Level 0} \hfill\hfill {Level 3} \hfill\hfill {Level 4} \hfill\hfill {Level 5} \hfill\hfill {Level 6} \hfill\hfill}
\end{center}
\vspace{-10pt}
\caption{Images generated from layouts with mixed precision levels. Number in the brackets indicate the precision levels. Regions of $0$-th levels are not shown. 
}
\vspace{-10pt}
\label{result_mix_levlec}
\end{figure*}

In the proposed framework, each semantic region can be assigned with an individual precision level to enable a more flexible control mechanism. In the paper, we have shown the use case of mixing a $0$-th level style indicator with other semantic regions of high levels. Here, we demonstrate more use cases of mixed precision levels in Fig.~\ref{result_mix_levlec}. For each input layout, we sample five images to reflect the difference in the strength of control of different precision levels. We can see that the object shapes in regions of lower precision shows more variations across different samples (\eg the bears in the first row and the skateboards in the second row) whereas those with higher precision are closer to each other.

\section{Additional Samples with Varying Precision Levels}
\begin{figure*}[h]
\begin{center}
%\fbox{\rule{0pt}{1.5in} \rule{\linewidth}{0pt}}
% \vspace{-10pt}
\includegraphics[width=\linewidth]{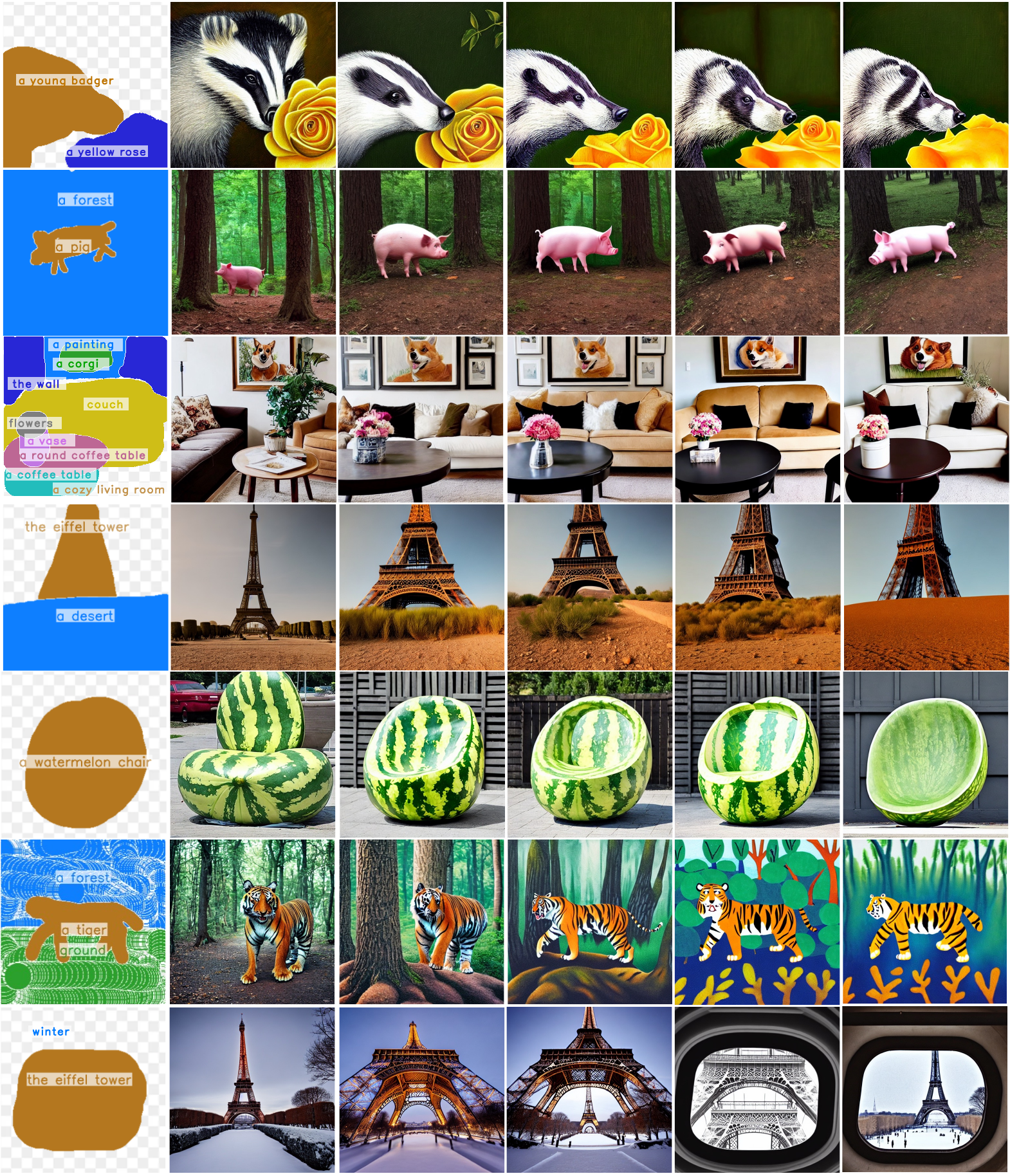}\\
\scriptsize{ \hfill {Input} \hfill\hfill {Level 0} \hfill\hfill {Level 3} \hfill\hfill {Level 4} \hfill\hfill {Level 5} \hfill\hfill {Level 6} \hfill\hfill}
\end{center}
\vspace{-10pt}
\caption{Results generated using different precision levels from the same layout and starting from the same noise. 
}
\vspace{-20pt}
\label{result_supp_levlec}
\end{figure*}

Fig.~\ref{result_supp_levlec} shows additional samples generated from layouts at different precision levels. When a coarse layout is assigned with a high precision leve, the model will still try to fit in the specified shapes by cartoonizing the images (\eg the last two results of the sixth row) or generating extra structures (\eg the extra photo frames in the last example). 

\section{Additional Analysis}
As indicated in Sec~4.2 of the paper, our method can generate images from bounding box layouts using the coarsest precision level. Here we report the quantitative comparison results on the COCO-Stuff validation set against the existing bounding box based methods in Table~\ref{table_box2im}. We process the data follow the settings of~\cite{sun2021learning,zhao2019image,johnson2018image}. Our method can achieve comparable FID to existing methods without being trained on bounding box layout, and outperforms them after being finetuned on COCO-Stuff training set with bounding box layout. 
\begin{table}[h]
\caption{\small Quantitative comparison with bounding box based layout-to-image synthesis models on COCO-Stuff. }
% \vspace{-5pt}
\begin{center}
\label{table_box2im}
% \resizebox{\columnwidth}{!}{
% }
\begin{tabular}{cccc}
\hline
 & Method & FID $\downarrow$ & zero-shot FID$\downarrow$\\
\hline
 \multirow{4}{*}{$64\times64$} & LostGAN-V1~\cite{sun2019image} & 34.31 & - \\
& OC-GAN~\cite{sylvain2021object} & 33.1 & - \\
& Layout2Im~\cite{zhao2019image} & 44.19 &  - \\
& Ours & 30.27 & 38.25 \\ 
% Ours &- & 9.47 \\ 
\hline
 \multirow{4}{*}{$256\times256$} & LostGAN-V2~\cite{sun2021learning} & 42.55 & - \\
& OC-GAN~\cite{sylvain2021object} & 41.65 & - \\
& LDM~\cite{rombach2022high} & 40.91 &  - \\
& Ours & 35.69 & 41.74 \\ 
% Ours &- & 9.47 \\ 
\hline
\end{tabular}
% \vspace{-10pt}
\end{center}
\end{table}

At inference time, we apply the classifier-free guidance based on the unconditional estimates from an empty text feature pyramid of the repeating $\bm{f}(\oslash)$. Here we explore the use of a partially empty text feature pyramid,~\ie only apply the classifier-free guidance with selected levels set to $\bm{f}(\oslash)$. The results are reported in Table~\ref{table_msguide}. Using classifier-free guidance with all levels yields the best performance (The last row of Table~\ref{table_msguide}). 
\begin{table}[h]
% \setlength{\tabcolsep}{1.5pt}
% \vspace{-10pt}
\caption{\small Results with classifier-free guidance applied at selected levels of the text feature pyramid. \textit{Level} indicates which levels to drop out when computing the unconditional estimates.}
% \vspace{-5pt}
\begin{center}
\label{table_msguide}
% \resizebox{\columnwidth}{!}{
\begin{tabular}{ccccc}
\hline
Level & CLIP Score $\uparrow$ & SS Score $\uparrow$\\
\hline
6 &.230 & .708\\
5,6 &.229& .714\\
4,5,6 &.229 & .718\\
3,4,5,6 & .229 & .733\\
0,3,4,5,6 & .261 & .736 \\
\hline
\end{tabular}
% }
% \vspace{-10pt}
\end{center}
\end{table}

We use the embedding of the concatenation of all regional text descriptions as a $1\times 1$ feature map at the $0$-th level of the text feature pyramid. We found the overall performance is not sensitive to the concatenation order, as indicated in Table~\ref{table_segperm}. 
\begin{table}[h]
\caption{\small Results with different concatenation orders. Ours: the class labels of different regions are concatenated according to the index in the COCO-Stuff dataset. Ours-permute: the class labels are concatenated in random order. }
% \vspace{-10pt}
\begin{center}
\label{table_segperm}
% \resizebox{\columnwidth}{!}{
\begin{tabular}{ccc}
\hline
Method & FID $\downarrow$ & mIOU$\uparrow$ \\
\hline
Ours &17.20 &23.01\\
Ours-permute & 17.21 & 22.76\\
\hline
\end{tabular}
% }
% \vspace{-10pt}
\end{center}
\end{table}

\section{Implementation Details}
\begin{figure*}[h]
\begin{center}
%\fbox{\rule{0pt}{1.5in} \rule{\linewidth}{0pt}}
% \vspace{-10pt}
\includegraphics[width=.6\linewidth]{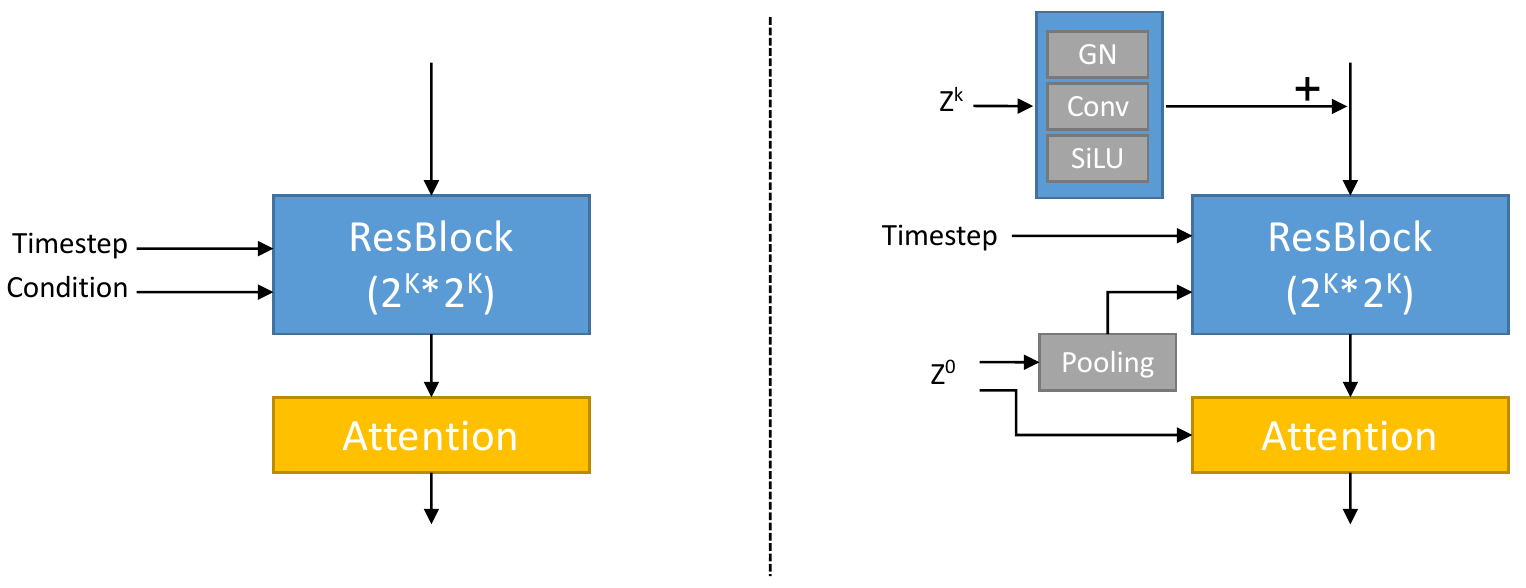}\\
% \scriptsize{\hfill {(a)} \hfill\hfill {(b)} \hfill\hfill {(c)} \hfill\hfill {(d)} \hfill\hfill {(e)} \hfill\hfill}
\end{center}
\vspace{-10pt}
\caption{The architecture of a block in the original UNet in~\cite{dhariwal2021diffusion} (left) and our modified UNet (right). GN: Group Normalization; SiLU: Sigmoid Linear Unit. 
}
\label{arch_layer}
\end{figure*}

Following the practices in~\cite{saharia2022photorealistic,rombach2022high,nichol2021glide}, we adopt the UNet architecture in~\cite{dhariwal2021diffusion} for the $64\times 64$ diffusion models. For the super-resolution model in experiments with the pixel-space diffusion, we use the same architecture as in~\cite{dhariwal2021diffusion}. We use the decoder from ~\cite{rombach2022high} for the latent-space diffusion model. 
The hyper-parameters for the $64\times 64$ model are summarized as follows.  
\small
\begin{verbatim}
    image_size: 64
    in_channels: 4 #3 for the pixel-space diffusion
    out_channels: 4 #3 for the pixel-space diffusion
    model_channels: 320 #192 for the pixel-space diffusion
    attention_resolutions: [ 32, 16, 8 ]
    num_res_blocks: 2
    channel_mult: [ 1, 2, 4, 4 ] #[1, 2, 3, 4] for the pixel-space diffusion
\end{verbatim}
\normalsize
Fig.~5 in the paper illustrates the overall architecture of the modified UNet. Here we show the architecture of each block in Fig.~\ref{arch_layer}. The original architecture in~\cite{dhariwal2021diffusion} is shown on the left side as a reference. 

We use the cosine noise schedule for diffusion. We uniformly sample from $1000$ diffusion steps in training. At inference time, we use the PLMS sampler~\cite{liu2022pseudo} and sample for $100$ diffusion steps. We set the classifier-free guidance scale to $3$ for evaluation in the text-to-image generation setting and $2$ for segmentation-to-image generation. The samples for visualization are generated with the guidance scale set to $7.5$. The precision level is set to $3$ by default if not specified. It takes $32$ seconds to generate a batch of $16$ samples on a A100 GPU. 

We annotate the layouts for training using text-based object  detection~\cite{li2021grounded} and segmentation~\cite{he2017mask}. More specifically, for an image with a caption, we use \cite{li2021grounded} to generate bounding boxes for entities that appear in the caption. Then we compute the similarity of each entity to the categories in COCO. 
We use the mask estimation branch of~\cite{li2021grounded} to obtain a segmentation mask if the maximum similarity is over $0.3$ and keep the mask of the bounding box otherwise.

\section{Limitations and Future Work}
\begin{figure*}[h]
\begin{center}
%\fbox{\rule{0pt}{1.5in} \rule{\linewidth}{0pt}}
% \vspace{-10pt}
\includegraphics[width=\linewidth]{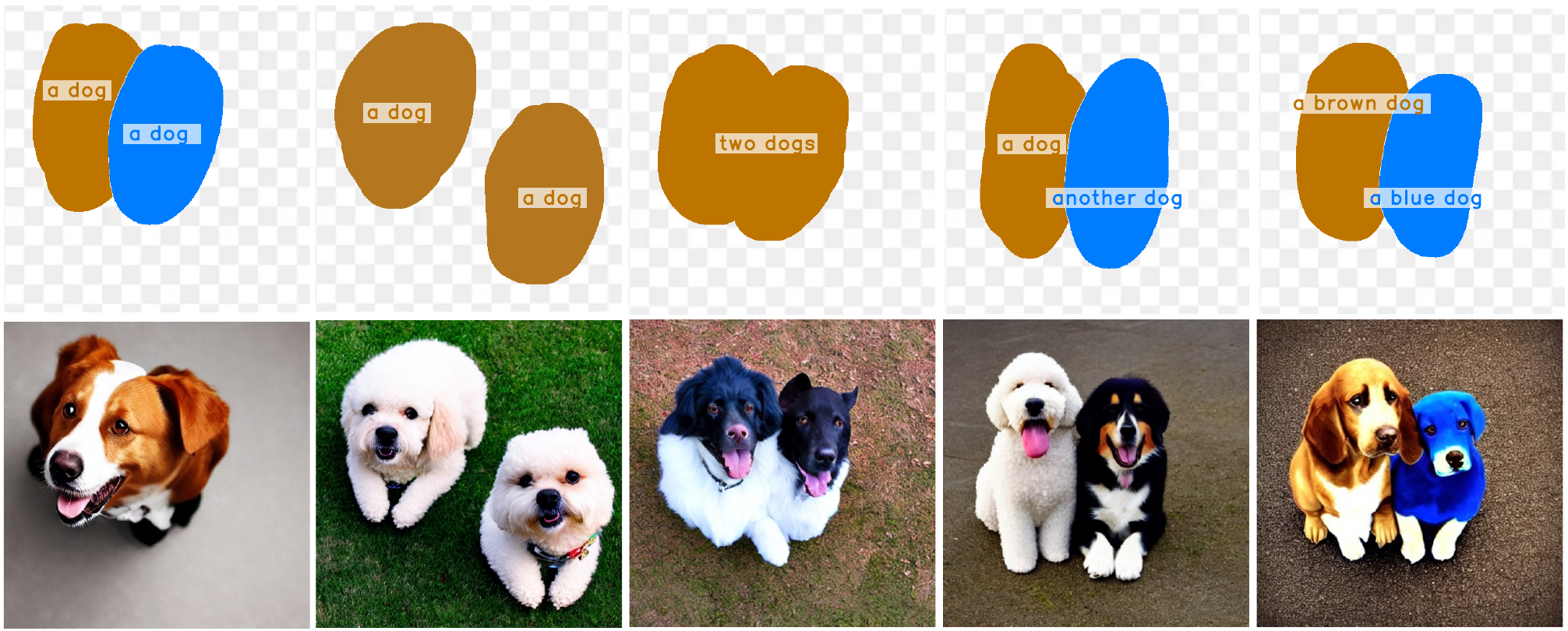}\\
\scriptsize{\hfill {(a)} \hfill\hfill {(b)} \hfill\hfill {(c)} \hfill\hfill {(d)} \hfill\hfill {(e)} \hfill\hfill}
\end{center}
\vspace{-10pt}
\caption{The text feature maps are not instance-aware. We need to specify the number of instances or attributes to separate instances of the same semantic category. 
}
\vspace{-20pt}
\label{result_limit_inst}
\end{figure*}

Since the text feature maps are not instance-aware, different instances of the same category or with the same text descriptions correspond to the same features. Therefore, the model cannot separate adjacent instances if their text descriptions are the same, as shown in Fig.~\ref{result_limit_inst}~(a). There are several ways to generate multiple instances with our model: 
\begin{itemize}[nosep]
\item Clearly define the instance boundary in the layout (Fig.~\ref{result_limit_inst}~(b)).
\item Specify the number of instances (Fig.~\ref{result_limit_inst}~(c)).
\item Emphasize multiple instances in text descriptions (Fig.~\ref{result_limit_inst}~(d)). 
\item Specify the attributes of different  (Fig.~\ref{result_limit_inst}~(e)).
\end{itemize}
However, these are still not very intuitive and require additional user effort. 
It can be a direction for future work to incorporate instance information in the proposed framework. 

\twocolumn

{\small
\bibliographystyle{ieee_fullname}
\bibliography{egbib}
}

\end{document}